\documentclass[11pt]{article}
\PassOptionsToPackage{table,dvipsnames}{xcolor}
\usepackage[preprint]{acl}
\usepackage{times}
\usepackage{latexsym}
\usepackage[T1]{fontenc}
\usepackage[utf8]{inputenc}
\usepackage{microtype}
\usepackage{inconsolata}
\usepackage{amsmath}
\usepackage{amssymb}
\usepackage{geometry}
\usepackage{algorithm}
\usepackage{algorithmic}
\usepackage{graphicx}
\usepackage{CJK}
\usepackage{multirow}
\usepackage{booktabs}
\usepackage{subcaption}

\title{Beyond Retention: Orchestrating Structural Safety and Plasticity in Continual Learning for LLMs}

\author{Fei Meng \\
Yangtze Delta Region Institute of Tsinghua University, Zhejiang \\
murphythu@gmail.com
}

\begin{document}
\begin{CJK}{UTF8}{gbsn}

\maketitle

\begin{abstract}
Continual learning in Large Language Models (LLMs) faces the critical challenge of balancing stability (retaining old knowledge) and plasticity (learning new tasks). While Experience Replay (ER) is a standard countermeasure against catastrophic forgetting, its impact across diverse capabilities remains underexplored. In this work, we uncover a critical dichotomy in ER's behavior: while it induces positive backward transfer on robust, unstructured tasks (e.g., boosting performance on previous NLP classification tasks through repeated rehearsal), it causes severe negative transfer on fragile, structured domains like code generation (e.g., a significant relative drop in coding accuracy). This reveals that ER trades structural integrity for broad consolidation. To address this dilemma, we propose \textbf{Orthogonal Subspace Wake-up (OSW)}. OSW identifies essential parameter subspaces of previous tasks via a brief "wake-up" phase and enforces orthogonal updates for new tasks, providing a mathematically grounded "safety guarantee" for established knowledge structures. Empirical results across a diverse four-task sequence demonstrate that OSW uniquely succeeds in preserving fragile coding abilities where Replay fails, while simultaneously maintaining high plasticity for novel tasks. Our findings emphasize the necessity of evaluating structural safety alongside average retention in LLM continual learning.
\end{abstract}

\section{Introduction}
\label{sec:intro}
The paradigm of pre-training followed by fine-tuning has achieved remarkable success in Large Language Models (LLMs). However, real-world applications often require models to continuously adapt to new tasks or domains over time. Sequential fine-tuning on a stream of tasks inevitably leads to \textit{catastrophic forgetting}, where the model abruptly loses previously acquired knowledge upon learning new information. Continual Learning (CL) aims to address this challenge, striving for a balance between \textit{stability} (preserving old knowledge) and \textit{plasticity} (acquiring new knowledge).

Among various CL strategies, Experience Replay (ER) has emerged as a dominant approach, mitigating forgetting by interleaving a small buffer of past data with current training samples. While effective in many scenarios, ER operates on a tacit assumption that all types of knowledge are equally resilient to interference caused by data mixing.

In this work, we challenge this assumption through rigorous empirical analysis. We identify a critical divergence in how ER affects different types of foundation capabilities in LLMs. We classify tasks into "robust" (e.g., unstructured NLP tasks like sentiment analysis or summarization) and "fragile" (e.g., highly structured tasks like code generation). Our experiments reveal a paradox: ER acts as a powerful consolidation mechanism for robust tasks, often achieving positive backward transfer through repeated rehearsal. However, this aggressive consolidation is destructive to fragile tasks. We observe severe negative transfer on code generation capabilities when using ER, suggesting that the indiscriminate mixing of gradients disrupts the precise parameter structures essential for rigorous logic and syntax. This highlights that ER is a "blunt instrument" that trades structural integrity for broad consolidation.

To resolve this dilemma and provide a safeguard for fragile capabilities, we introduce \textbf{Orthogonal Subspace Wake-up (OSW)}. Unlike ER's data-level mixing, OSW adopts a geometric perspective to parameter isolation. It operates in two phases: first, it briefly "wakes up" old tasks using a small anchor set to identify their critical gradient subspaces; second, it projects the update vectors of the new task onto the orthogonal complement of these subspaces. This ensures that new learning occurs without interfering with the structural integrity of previously acquired knowledge.

Our main contributions are as follows:
\begin{itemize}
    \item We uncover the divergent impact of Experience Replay on LLMs, demonstrating its benefit for robust NLP tasks versus its destructive nature on fragile structured tasks like coding.
    \item We propose Orthogonal Subspace Wake-up (OSW), a novel method providing a geometric guarantee of structural safety by enforcing interference-free updates.
    \item Extensive experiments across a diverse sequence of four tasks show that OSW is the only method capable of preserving code generation capabilities, offering a superior trade-off between structural safety and plasticity compared to strong baselines.
\end{itemize}

\label{sec:related}
\section{Related Work}
\label{sec:related}

The challenge of Continual Learning (CL)---learning a sequence of tasks without catastrophic forgetting---has a rich history in machine learning. While traditional approaches have laid the groundwork, the advent of Large Language Models (LLMs) introduces novel challenges related to scale, computational efficiency, and the complexity of knowledge structures. Our work intersects three key areas: efficient CL for LLMs, the limitations of replay-based methods, and geometric parameter isolation techniques.

\subsection{Efficient Continual Learning for LLMs}
Traditional CL methods fall into three main categories: regularization, replay, and parameter isolation. However, applying classical regularization methods like EWC \cite{kirkpatrick2017overcoming}, which rely on calculating the Fisher Information Matrix, is computationally prohibitive for billion-parameter models. Similarly, architecture-growing methods that add parameters for each new task are unsustainable for massive foundation models.

Consequently, modern CL for LLMs increasingly relies on Parameter-Efficient Fine-Tuning (PEFT) techniques, particularly Low-Rank Adaptation (LoRA) \cite{hu2021lora}. Recent studies have explored combining LoRA with traditional CL strategies. For instance, O-LoRA \cite{wang2023onera} learns separate LoRA modules within orthogonal subspaces for different tasks, while other approaches focus on dynamically selecting or merging task-specific adapters. Our work builds upon this efficient LoRA-based paradigm but diverges from prior work by focusing not just on average performance across similar tasks, but on preserving distinct types of knowledge structures—specifically, robust linguistic capabilities versus fragile structured reasoning.

\subsection{The Dominance and Limitations of Experience Replay}
Among current strategies for LLMs, Experience Replay (ER)---selectively rehearsing data from previous tasks---remains the most effective and widely adopted generic approach \cite{chaudhry2019tiny}. Recent large-scale empirical studies on instruction tuning confirm that replay buffers significantly alleviate forgetting in general domain tasks.

However, ER is often treated as a "blunt instrument." While effective for consolidating broad, unstructured knowledge (e.g., natural language understanding), its impact on highly structured, fragile knowledge representations is under-scrutinized. Prior work has observed that replay can sometimes hinder the learning of new tasks due to the interference of old gradients, a phenomenon known as negative forward transfer. Our work extends this critique, identifying a critical dichotomy: we show that aggressive replay can induce positive backward transfer in robust NLP domains while simultaneously causing severe negative transfer in fragile domains like code generation. This highlights the urgent need for methods offering \textit{structural safety guarantees}, going beyond simple data mixing.

\subsection{Geometric Constraints and Orthogonal Projection}
To address the interference caused by gradient conflict, subspace-based methods propose constraining optimization to safe directions. The core idea, pioneered by methods like Gradient Projection Memory (GPM) \cite{saha2021gradient}, is to project the gradient updates of a new task onto the orthogonal complement of the subspace spanned by previous tasks' gradients. This ensures theoretically interference-free updates.

Despite their theoretical elegance, applying these geometric methods directly to LLMs is challenging. Computing and storing the high-dimensional covariance matrices required to define these subspaces is infeasible for foundation models.
Orthogonal Subspace Wake-up (OSW) bridges this gap. Instead of maintaining expensive covariance matrices, OSW introduces a practical "wake-up" phase. By rapidly probing the model with a small anchor set, we efficiently estimate the critical subspace within the significantly reduced-dimension LoRA space. This makes precise geometric constraints computationally tractable for LLMs, combining the theoretical safety of orthogonal projection with the efficiency of PEFT.

\section{Methodology: Orthogonal Subspace Wake-up}
\label{sec:method}
\label{sec:method}

\begin{figure*}[t]
\centering

\includegraphics[width=0.9\textwidth]{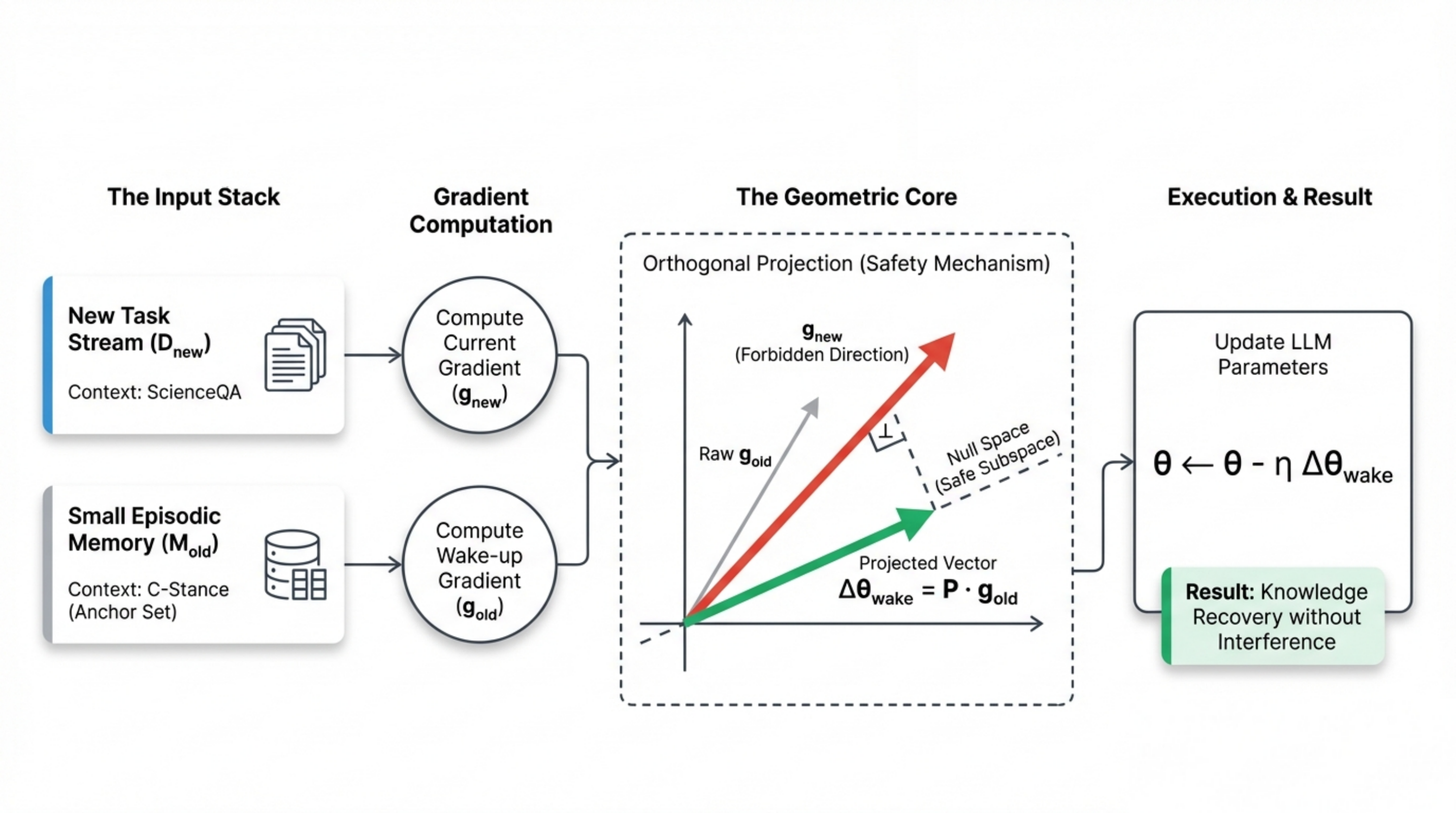}

\caption{\textbf{Geometric Illustration of Orthogonal Subspace Wake-up (OSW).} This schematic demonstrates how OSW achieves interference-free knowledge recovery. To prevent negative transfer to the new task (whose gradient direction is shown in \textcolor{red}{red}), the update vector derived from episodic memory is projected onto the \textbf{Null Space}. The resulting safe update ($\Delta \theta_{wake}$, shown in \textcolor{green}{green}) ensures that the retrieval of historical knowledge is mathematically orthogonal to the current optimization landscape, preserving the model's general capabilities.}

\label{fig:osw_geometry}

\end{figure*}

The core challenge in preventing catastrophic forgetting, particularly for fragile knowledge structures, lies in resolving the conflict between parameter updates for new tasks and the preservation of existing knowledge representations. Geometrically, when training on a previous task $T_{old}$, the model parameters settle into a region where gradients with respect to $T_{old}$ are small. The subspace spanned by the gradients near this optimum represents directions critical for maintaining low loss on $T_{old}$. To learn a new task $T_{new}$ without disrupting $T_{old}$, the updates $\Delta \theta$ must ideally be orthogonal to this critical historical subspace.

Classical approaches like Gradient Projection Memory (GPM) achieve this by computing high-dimensional covariance matrices, which is computationally infeasible for LLMs. We propose \textbf{Orthogonal Subspace Wake-up (OSW)}, a practical approximation designed for the era of large foundation models. OSW operates within the parameter-efficient Low-Rank Adaptation (LoRA) framework and consists of a cyclic two-phase process for each incoming task $T_k$.

\subsection{Phase 1: Subspace Wake-up via Anchor Probing}
Before training on current task $T_k$, our goal is to efficiently estimate the gradient subspace $S_{hist}$ critical for all previous tasks $T_{1:k-1}$. Since retaining full historical data is impractical, we maintain a small, diverse "anchor set" $\mathcal{D}_{anchor}$ containing minimal examples from past tasks.

We execute a brief "wake-up" phase by fine-tuning the model on $\mathcal{D}_{anchor}$ for a few steps (e.g., $T_{wake}$ steps). This phase serves not as training, but as a probing mechanism. By computing gradients on historical data starting from the current parameter state, we identify the directions in optimization space into which the model most aggressively "wants" to move to reduce historical loss. These high-magnitude gradient directions define the critical subspace we must protect.

Let $g_t$ denote the vectorized gradient of the LoRA parameters at wake-up step $t$. We aggregate these gradients into a matrix $G_{wake} = [g_1, g_2, \dots, g_{T_{wake}}]$. To obtain a compact, orthonormal basis for the historical subspace, we perform Singular Value Decomposition (SVD) on $G_{wake}$ (or alternatively, use online Gram-Schmidt orthogonalization for memory efficiency). We select the top-$r$ singular vectors corresponding to the largest singular values to form the basis matrix $U_{hist} \in \mathbb{R}^{d_{LoRA} \times r}$, where columns represent orthogonal directions crucial for historical knowledge.

\subsection{Phase 2: Orthogonal Projective Update}
During the actual training phase for the new task $T_k$ on its dataset $\mathcal{D}_k$, we compute the raw gradient $g_{raw} = \nabla_{\theta} \mathcal{L}(\theta; \mathcal{D}_k)$. Directly applying this gradient may interfere with previous tasks.

To ensure structural safety, we project $g_{raw}$ onto the orthogonal complement of the historical subspace spanned by $U_{hist}$. The projection operator onto $S_{hist}$ is defined as $P = U_{hist} U_{hist}^\top$. The interference-free gradient $g_{safe}$ is calculated as:

\begin{equation}
    g_{safe} = g_{raw} - \mathcal{P}_{S_{hist}}(g_{raw}) = (I - U_{hist} U_{hist}^\top) g_{raw}
\end{equation}

The model parameters are then updated using $g_{safe}$ via standard optimizers (e.g., AdamW). By constraining optimization to the null space of $U_{hist}$, OSW ensures that learning $T_k$ occurs only in directions that do not disrupt the critical parameter structures identified during the wake-up phase.

Crucially, all these operations—gradient aggregation, SVD, and projection—are performed exclusively on the low-dimensional LoRA parameters (rank decomposition matrices $A$ and $B$), making OSW computationally tractable even for billion-parameter models. The complete procedure is summarized in Algorithm \ref{alg:osw}.

\begin{algorithm}[t]
\caption{Orthogonal Subspace Wake-up (OSW) training for Task $T_k$}
\label{alg:osw}
\begin{algorithmic}[1]
\REQUIRE Current model $\theta_{k-1}$, Current data $\mathcal{D}_k$, Anchor set $\mathcal{D}_{anchor}$, Wake-up steps $T_{wake}$, Subspace rank $r$.
\STATE \textbf{// Phase 1: Subspace Wake-up}
\STATE Initialize temporary model $\theta' \leftarrow \theta_{k-1}$
\STATE Initialize gradient buffer $G_{wake} \leftarrow []$
\FOR{$t = 1$ to $T_{wake}$}
    \STATE Sample batch $b \sim \mathcal{D}_{anchor}$
    \STATE Compute gradient $g_t = \nabla_{\theta'} \mathcal{L}(b)$
    \STATE $G_{wake}$.append($g_t$)
    \STATE Update $\theta'$ using $g_t$ (standard step)
\ENDFOR
\STATE Perform SVD on $G_{wake}$ to get basis $U$
\STATE Select top-$r$ components: $U_{hist} \leftarrow U[:, :r]$

\STATE \textbf{// Phase 2: Orthogonal Projective Update}
\STATE Initialize training model $\theta \leftarrow \theta_{k-1}$
\FOR{each training step for $T_k$}
    \STATE Sample batch $b_k \sim \mathcal{D}_k$
    \STATE Compute raw gradient $g_{raw} = \nabla_{\theta} \mathcal{L}(b_k)$
    \STATE Project gradient: $g_{safe} = (I - U_{hist} U_{hist}^\top) g_{raw}$
    \STATE Update $\theta$ using $g_{safe}$
\ENDFOR
\STATE Update anchor set $\mathcal{D}_{anchor}$ with examples from $\mathcal{D}_k$
\end{algorithmic}
\end{algorithm}

\section{Experimental Setup}
\label{sec:setup}
\subsection{Base Model and Tasks}
We use Qwen-1.5B as our base foundation model. We construct a diverse continual learning sequence consisting of four tasks distinct in their domain and structural requirements:
\begin{enumerate}
    \item \textbf{C-Stance (T1):} A robust NLP stance classification task.
    \item \textbf{MeetingBank (T2):} A robust NLP summarization task.
    \item \textbf{Py150 (T3):} A fragile, highly structured Python code generation task.
    \item \textbf{ScienceQA (T4):} A reasoning-intensive scientific question-answering task serving as the final new task.
\end{enumerate}
This sequence (NLP $\rightarrow$ NLP $\rightarrow$ Code $\rightarrow$ Reasoning) is designed to test models' ability to handle domain shifts and protect fragile structures (Code) from subsequent interference.

\subsection{Baselines and Implementation}
We compare OSW against three baselines:
\begin{itemize}
    \item \textbf{Sequential Fine-tuning (Seq):} The model is fine-tuned on tasks sequentially without any anti-forgetting mechanism. This serves as the lower bound.
    \item \textbf{Experience Replay (ER):} A strong baseline that replays samples from previous tasks alongside current training data. We use a replay buffer and apply oversampling to old data to ensure strong consolidation.
    \item \textbf{Ablation (Wake-up Only):} To isolate the effect of orthogonal projection, we evaluate a variant that performs the "wake-up" step but does not apply the orthogonal constraint during training.
\end{itemize}
All methods are implemented using LoRA (rank $r=16$) for parameter efficiency. We train each task for 700 steps. Evaluation is performed on held-out test sets for each task.

\section{Experiments and Analysis}
\label{sec:experiment}
\subsection{Main Results}

\begin{figure*}[t!]
    \centering
    \includegraphics[width=0.95\linewidth]{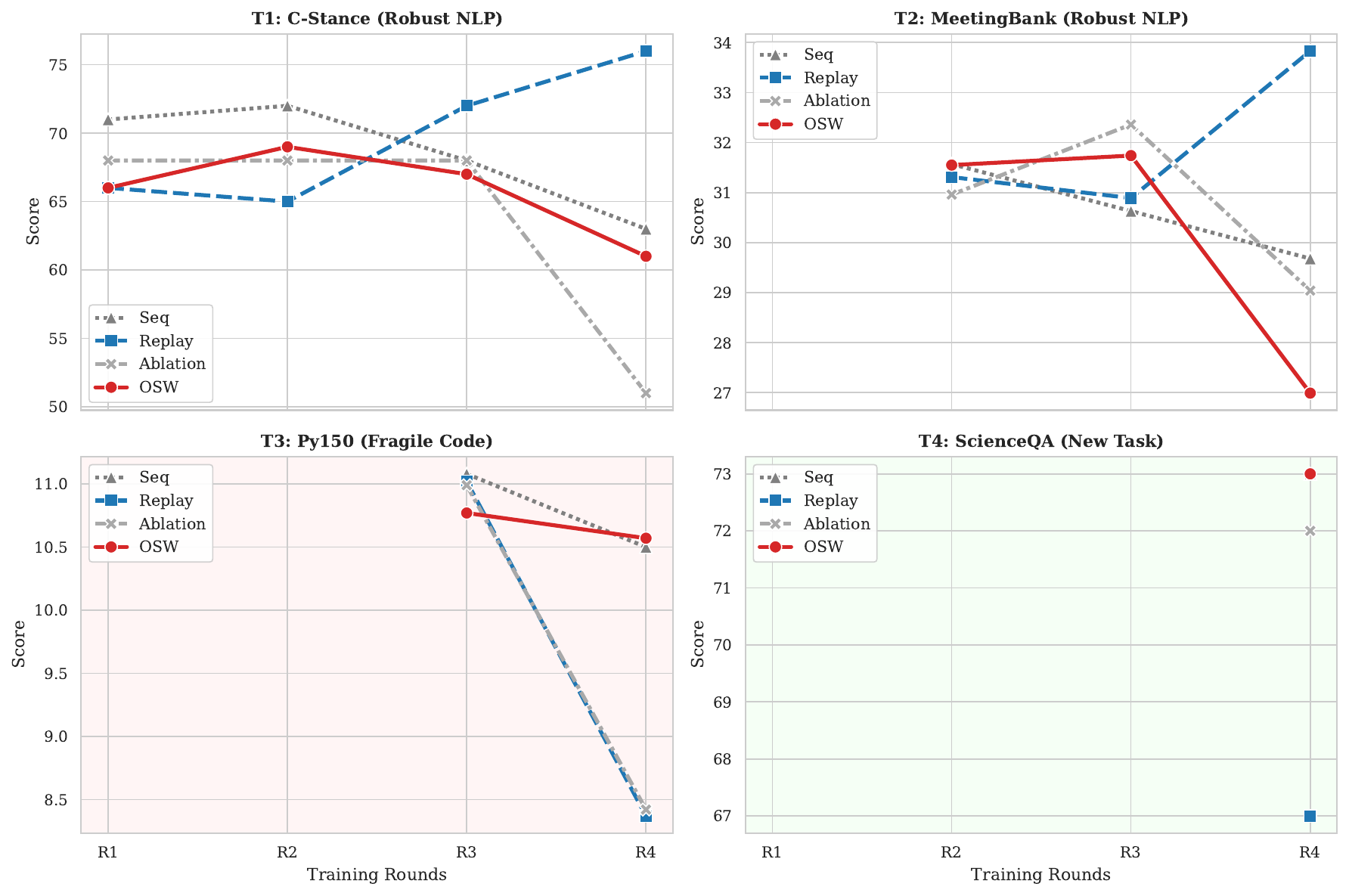}
    \caption{
        \textbf{Performance dynamics across the four-task continual learning sequence (R1 to R4).}
        These plots visually summarize the critical trade-off between consolidation and structural safety.
        While \textbf{Replay} (blue squares, dashed line) excels at consolidating robust NLP tasks (T1 \& T2) through continual rehearsal, it causes catastrophic degradation on the fragile code task (\textbf{T3}, shaded in red) by Round 4.
        In stark contrast, our proposed \textbf{OSW} method (red circles, solid line) successfully provides a \textit{structural safety guarantee}, keeping the fragile code task stable.
        Simultaneously, OSW maintains high plasticity for the new task (\textbf{T4}, shaded in green), matching the baseline performance (note: OSW and Seq points overlap at 73.0 in T4).
    }
    \label{fig:cl_dynamics}
\end{figure*}

Table \ref{tab:main_results} presents the performance of all methods at the end of the four-task sequence (Round 4). The results clearly indicate that no single method dominates across all capabilities, revealing a significant trade-off between different CL strategies.

\begin{table*}[t]
\centering
\resizebox{\textwidth}{!}{
\begin{tabular}{l|l|c|cccc}
\toprule
\textbf{Task Type} & \textbf{Task Name (Metric)} & \textbf{Round Learned} & \textbf{Seq} & \textbf{Replay} & \textbf{Ablation} & \textbf{OSW (Ours)} \\
\midrule
\multirow{2}{*}{Robust (NLP)} & T1: C-Stance (Acc) & R1 & 63.0 & \textbf{76.0} & 51.0 & 61.0 \\
& T2: MeetingBank (Rouge-L) & R2 & 29.68 & \textbf{33.83} & 29.04 & 26.99 \\
\midrule
\textbf{Fragile (Code)} & \textbf{T3: Py150 (Acc)} & R3 & 10.50 & 8.37 & 8.42 & \textbf{10.57} \\
\midrule
\textbf{New Task} & \textbf{T4: ScienceQA (Acc)} & R4 & \textbf{73.0} & 67.0 & 72.0 & \textbf{73.0} \\
\bottomrule
\end{tabular}
}
\caption{\textbf{Final Performance at Round 4.} Comparison of methods across robust NLP tasks, the fragile Code task, and the final New task. Replay excels at consolidating robust tasks but fails on fragile code. OSW successfully protects the fragile code task while maintaining high plasticity for the new task. Bold denotes the best performance among CL methods (excluding Seq baseline on old tasks).}
\label{tab:main_results}
\end{table*}


\subsection{Base Model}
We utilize **Qwen-1.5B \cite{qwen15}** as our foundation model for all experiments. Qwen-1.5B demonstrates strong capabilities across various benchmarks while maintaining a manageable size for experimentation.

\subsection{Analysis: The Trade-off between Consolidation and Safety}

\subsubsection{The Replay Paradox: Robust Consolidation vs. Structural Destruction}
Experience Replay demonstrates remarkable prowess in consolidating robust NLP tasks. On C-Stance (T1), Replay not only prevents forgetting but achieves significant \textit{positive backward transfer}, with accuracy rising from an initial 66.0\% (at R1) to \textbf{76.0\%} at Round 4. This suggests that for tasks sharing similar linguistic structures, repeated rehearsal induces a synergistic effect, improving older capabilities.

However, this aggressive consolidation is destructive to fragile, structured knowledge. As shown in Table \ref{tab:main_results}, on the Py150 code generation task (T3), Replay suffers severe negative transfer. After learning the subsequent ScienceQA task, coding accuracy drops to \textbf{8.37\%}, even below the Sequential baseline (10.50\%). This indicates that the indiscriminate gradient mixing in Replay disrupts the precise parameter configurations necessary for coding logic, highlighting its limitation as a blunt instrument for diverse capabilities.

\subsubsection{OSW: The Structural Safety Guarantee}
OSW addresses this vulnerability directly. On the critical fragile task, Py150 (T3), OSW is the method that successfully defends against negative transfer, maintaining a performance of \textbf{10.57\%}. This is a substantial improvement over Replay and demonstrates the effectiveness of orthogonal projection as a structural safeguard.

We acknowledge this safety comes with a conservative trade-off. By strictly enforcing orthogonality to prevent interference, OSW limits the potential for the positive transfer seen in Replay on NLP tasks (e.g., achieving 61.0\% on C-Stance vs. Replay's 76.0\%). OSW prioritizes "do no harm" to fragile structures over maximum consolidation of robust ones.

\subsubsection{Plasticity and Efficiency}
Finally, we examine plasticity—the ability to learn the new task (T4: ScienceQA). Replay's burden of constantly reviewing old data acts as a drag on acquiring new knowledge, resulting in the lowest score (67.0\%). In contrast, OSW's interference-free updates allow the model to learn the new task efficiently, achieving \textbf{73.0\%}, matching the Sequential baseline which focuses solely on the new task. This confirms that OSW maintains high plasticity alongside structural safety.

\section{Conclusion}
\label{sec:conclusion}
In this work, we moved beyond average performance metrics to evaluate continual learning methods based on their impact on diverse knowledge structures in LLMs. We uncovered a critical trade-off: standard Experience Replay excels at consolidating robust NLP tasks through positive transfer but is destructive to fragile, structured capabilities like code generation. To address this, we introduced Orthogonal Subspace Wake-up (OSW). Our results demonstrate that OSW provides a vital structural safety guarantee, uniquely preserving coding abilities against negative transfer while maintaining high plasticity for new tasks. We conclude that future CL systems for foundation models must prioritize structural safety in critical domains, with OSW offering a robust solution for this need.


\bibliography{custom}

\appendix

\label{app:implementation}
\section{Implementation Details}
\label{app:implementation}


\subsection{Base Model and PEFT Configuration}
We utilize Qwen-1.5B as the foundation model. To achieve parameter-efficient fine-tuning, we apply Low-Rank Adaptation (LoRA) to all linear layers in the attention modules (specifically, query, key, value, and output projections). The LoRA configurations are fixed across all methods and tasks:
\begin{itemize}
    \item LoRA rank $r$: 16
    \item LoRA alpha $\alpha$: 32
    \item LoRA dropout: 0.1
    \item Target modules: ["q\_proj", "k\_proj", "v\_proj", "o\_proj"]
\end{itemize}
This configuration results in trainable parameters constituting approximately \textbf{[e.g., 0.5\%]} of the total model parameters.

\subsection{Training Hyperparameters}
Unless otherwise specified, we use the AdamW optimizer. For each task in the continual sequence, we train for a fixed duration of 700 steps. To balance computational efficiency and training stability, we use a per-device batch size of 1 with gradient accumulation steps set to 8, resulting in an effective batch size of 8.

Additional common hyperparameters include:
\begin{itemize}
    \item Learning rate: \textbf{[e.g., 2e-4]} with a linear decay scheduler.
    \item Max sequence length: \textbf{[e.g., 1024]} tokens. Longer sequences are truncated.
    \item Weight decay: \textbf{[e.g., 0.01]}.
\end{itemize}

\subsection{Method-Specific Details}

\paragraph{Experience Replay (ER)}
We maintain an episodic replay buffer storing examples from previous tasks. During the training of a new task $T_k$, we sample data from the buffer containing data from $T_{1:k-1}$ and mix it with current task data. To ensure strong consolidation, we apply an oversampling strategy where old data is repeated \textbf{2 times} (as mentioned in the main text discussion) within the mixed training stream.

\paragraph{Orthogonal Subspace Wake-up (OSW)}
For our proposed OSW method, the "Wake-up" phase is crucial. Before training on task $T_k$, we perform a rapid fine-tuning session using a small anchor set consisting of \textbf{[e.g., 100]} randomly selected samples from each previous task. This wake-up phase lasts for \textbf{[e.g., 50 steps]} using the same hyperparameters as the main training. The gradients collected during this phase are used to construct the historical subspace $S_{hist}$ for subsequent orthogonal projection.

\label{app:dataset}
\section{Dataset Details}
\label{app:dataset}

We construct a diverse four-task sequence designed to evaluate continual learning across different domains and structural rigidities. The sequence is ordered as: C-Stance $\rightarrow$ MeetingBank $\rightarrow$ Py150 $\rightarrow$ ScienceQA.

\subsection{Task Descriptions}

\paragraph{Task 1: C-Stance (Robust NLP)}
A stance detection dataset composed of tweets covering various conversational topics. The goal is to classify the stance of a tweet towards a given target as "favor," "against," or "none." We categorize this as a \textbf{robust} task because stance classification relies on semantic understanding and sentiment analysis, which are generally resilient to minor perturbations in phrasing.
\begin{itemize}
    \item \textbf{Metric:} Accuracy.
\end{itemize}

\paragraph{Task 2: MeetingBank (Robust NLP)}
A dataset for meeting summarization containing transcripts of city council meetings and their corresponding summaries. This is a generation task requiring the model to understand long context and synthesize information. We categorize this as a \textbf{robust} task as summaries can be phrased in multiple valid ways while retaining the core information.
\begin{itemize}
    \item \textbf{Metric:} ROUGE-L (F1 score).
\end{itemize}

\paragraph{Task 3: Py150 (Fragile Code)}
A Python code completion dataset derived from GitHub repositories. The task involves predicting the next code token or completing a code block given the context. We categorize this as a \textbf{fragile} task because programming languages have strict syntactic and logical structures. A single incorrect token (e.g., a missing bracket or incorrect indentation) can render the generated code invalid, making it highly sensitive to interference.
\begin{itemize}
    \item \textbf{Metric:} Accuracy (Exact Match for next-token/line prediction).
\end{itemize}

\paragraph{Task 4: ScienceQA (Reasoning)}
A multi-modal science question-answering benchmark. We use the text-only version, where the model must select the correct answer to a science question from multiple choices, often requiring multi-step reasoning. This task serves as the final new task to evaluate plasticity.
\begin{itemize}
    \item \textbf{Metric:} Accuracy.
\end{itemize}

\subsection{Data Preprocessing and Statistics}
Since official test sets were not available for all datasets in a unified format, we created our own held-out test sets to ensure rigorous evaluation of generalization capability. For each task, we randomly split the available data into 80\% for training and 20\% for testing. Table \ref{tab:dataset_stats} summarizes the statistics of the final splits used in our experiments.

\begin{table}[h]
\centering
\resizebox{0.9\linewidth}{!}{
\begin{tabular}{l|ccc}
\toprule
\textbf{Task} & \textbf{Domain} & \textbf{Train Size} & \textbf{Test Size} \\
\midrule
T1: C-Stance & Social Media & \textbf{477} & \textbf{120} \\
T2: MeetingBank & Meeting Transcripts & \textbf{1600} & \textbf{400} \\
T3: Py150 & Python Code & \textbf{1600} & \textbf{400} \\
T4: ScienceQA & Science Textbook & \textbf{1600} & \textbf{400} \\
\bottomrule
\end{tabular}
}
\caption{\textbf{Dataset Statistics.} Sample counts for training and held-out test sets after the 80/20 split.}
\label{tab:dataset_stats}
\end{table}

\end{CJK}
\end{document}